\documentclass[10pt,letterpaper]{article}

\usepackage{cogsci}

\usepackage{graphicx}
\graphicspath{ {./figs/} } 

\usepackage{amsmath}

\cogscifinalcopy 

\usepackage{pslatex}
\usepackage{apacite}
\usepackage{float} 

\usepackage{fixltx2e} 

\title{Habits of Mind: Reusing Action Sequences for Efficient Planning}
 
\author{{\large \bf Noémi Éltető$^1$ (noemi.elteto@tuebingen.mpg.de)} \& {\large \bf Peter Dayan$^{1,2}$ (dayan@tue.mpg.de)} \\[-15pt] \AND
  $^1$Computational Neuroscience, MPI for Biological Cybernetics \quad 
  $^2$University of T\"ubingen}

\begin{document}

\maketitle

\begin{abstract}

When we exercise sequences of actions, their execution becomes more fluent and precise. Here, we consider the possibility that exercised action sequences can also be used to make planning faster and more accurate by focusing expansion of the search tree on paths that have been frequently used in the past, and by reducing deep planning problems to shallow ones via multi-step jumps in the tree. To capture such sequences, we use a flexible Bayesian action chunking mechanism which  finds and exploits statistically reliable structure at different scales. This gives rise to shorter or longer routines that can be embedded into a Monte-Carlo tree search planner. We show the benefits of  this scheme using a physical construction task patterned after tangrams.

\textbf{Keywords:} 
Planning; Sequence models; Bayesian nonparametrics; Action chunking;
\end{abstract}

\section{Introduction}

The \textit{law of exercise} holds that animals have a tendency to repeat or perseverate on frequent choices, even when this reduces reward \cite{thorndike1898animal, miller2019habits}. This leveraging  of structure in behaviour complements the leveraging of structure in the environment.  \citeA{gershman2020origin} suggested that perseveration arises from a trade-off between reward maximization and policy complexity, with repetition making policies more compact. This theory was focused on isolated actions;  however, animals also pick up on \textit{sequences}  of action. The action chunking process that underlies skill or habit learning \cite{dhawale2021basal} has also been observed in multi-step decision-making tasks, for instance with rats repeating whole trajectories of choices \cite{dezfouli2012habits}. A hallmark of such action chunks is that they are performed quickly and uninterrupted by state assessment, that is, open-loop.

The law of exercise may apply not only to the \textit{execution} of actions in multi-step problems, but also to the \textit{planning} thereof. \citeA{huys2015interplay} tested how human participants simplified a deep planning problem. One of the three heuristics they found was stochastic memoization, in which  participants became increasingly prone to repeat previous action paths in their entirety or chunks of them, even though these might not be optimal. In their interpretation, this marks a transfer from flexible but costly computation to the less flexible but cheaper reliance on past experience. A similar observation was made in a task where participants planned and constructed complex structures that were required to be stable in a virtual environment simulating rigid-body physics \cite{mccarthy2020learning}. During training, participants' construction procedures became increasingly stereotyped.

The trouble with sequences of actions is that their number tends to grow exponentially with the length of the sequence. This makes methods that try to be comprehensive \cite{riegler2021embodiment} computationally infeasible. Indeed, we know of no scalable or cognitively plausible model of action chunk reuse in planning. Here, in keeping with a general trend towards internalized decision-making  \cite{dayan2012set}, we examine the benefits of using a non-parametric  Bayesian sequence model that can infer sparse chunks of various lengths, depending on the statistics of the 
sequences executed in past episodes. We use this model to 
augment Monte-Carlo Tree Search (MCTS) in two ways: using single actions that it predicts as being likely, to bias one-step tree expansions, and  proposing popular multi-step expansions.  We test our model on a physical construction task called the Sticky Tangram.

\section{MCTS-with-HABITS}

\begin{figure}[!tb]
\begin{center}
\includegraphics[scale=0.8]{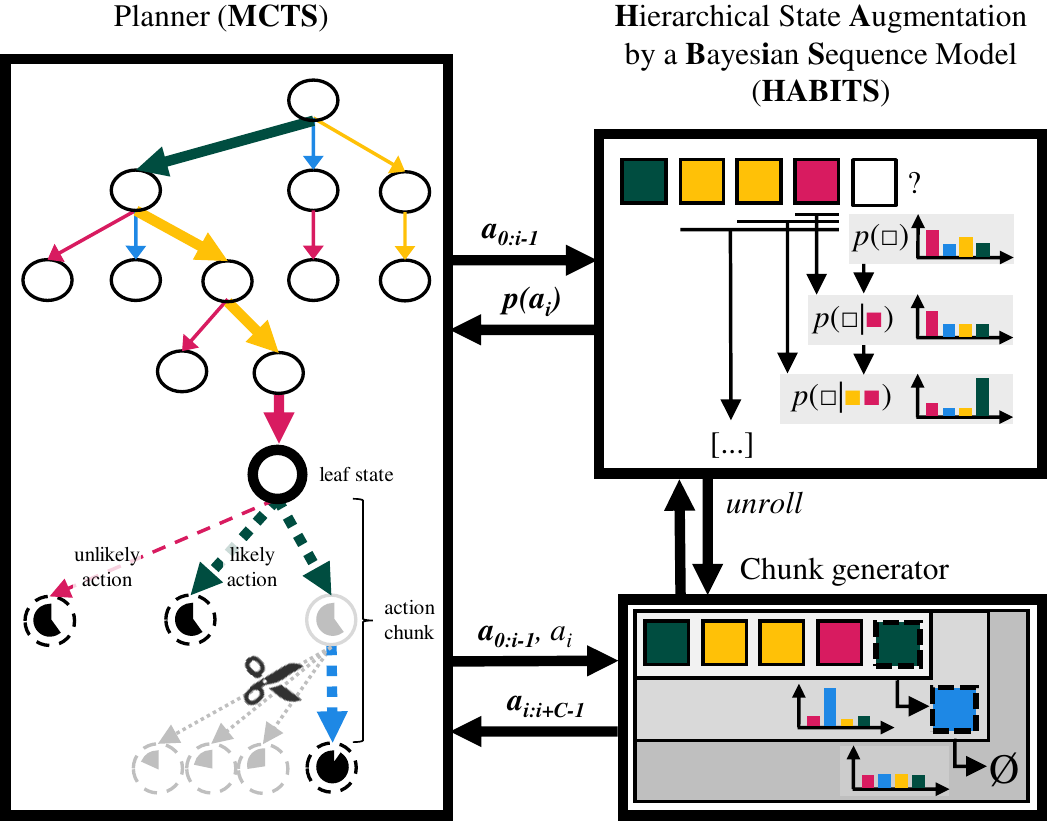}
\end{center}
\caption{Schematic of the MCTS-with-HABITS model. The planner builds a search tree with nodes representing states and edges representing different possible actions, marked by colors (left box). The planner traverses the tree by choosing actions that ultimately lead to more simulated wins and that are more predictable by the HABITS sequence model (upper right box). At the leaf state (marked by the bold node) several potential actions (marked by dashed edges) are considered that lead to states (marked by dashed nodes) that have not been added to the tree yet. The \textit{red} and the \textit{green} actions are primitive actions whose winning values (marked as pies) are the same. Yet, the \textit{green} action has a higher likelihood under the sequence model, when conditioned on the action trace from the root to the leaf node. Therefore, the \textit{green} action will be preferred over \textit{red}. The \textit{green} action is also predictably followed, given the context of the action trace, by the \textit{blue} action. Therefore, the chunk generator (lower right box) appends the \textit{blue} action to the \textit{green} one. Conditioned on the previous action trace \textit{and} the \textit{blue} action, no further action is strongly predicted and the chunk generation is terminated.
Thus we extend the available action set by the \textit{green-blue} chunk. Note that the primitive \textit{green} action, from which we unrolled a chunk, is also kept as an alternative. If the action chunk is selected by the planner then it jumps over the state marked by a grey node, not considering the available actions from that state, also marked by grey. This effective stunting results in a different value estimate for the chunk \textit{green}-\textit{blue} (marked by the larger pie in the node) compared to the primitive action \textit{green}; in this example, the chunk will be preferred by the tree policy.} 
\label{model_figure}
\end{figure}

To examine the benefits of biasing planning towards previously learned action sequences, we used a Bayesian infinite sequence model (a hierarchical Dirichlet process) to augment Monte Carlo Tree Search with open-loop, multi-action, node expansions. We call the new planning algorithm MCTS-with-HABITS (Figure~\ref{model_figure}).

Consider the search tree in Figure~\ref{model_figure}, representing a deterministic planning problem in which nodes correspond to states and the edges correspond to actions. In Monte Carlo Tree Search (MCTS; \citeA{kocsis2006bandit}), the tree is traversed from the root node to a leaf by repeated action choices based on their simulated expected value. This decision mechanism is called the tree policy. A leaf is any node that has a potential child whose value has not been estimated. Once a leaf is reached, it is expanded by considering the actions that are available in that state and choosing one of them. Then, the value of the node is estimated by simulating a random rollout using a uniform random policy. The reward of the rollout (in our case of a puzzle, $1$ for a correct solution; $0$ otherwise) is backpropagated to the selected node and its ancestors. The four ingredients -- selection, expansion, simulation, and backpropagation -- constitute one step of the MCTS algorithm. Since it estimates the value of nodes via simulation, MCTS focuses the search on promising paths without the need for handcrafted distance heuristics  as in A*. In order to balance this exploitative focus (from the fraction $\frac{w_t}{n_t}$ of wins $w_t$ from $n_t$ tries from a node on the $t^{\text{th}}$ MCTS step) with exploration of potentially even better states, a bonus is added to actions leading to lesser tried nodes ($\sqrt{\frac{ln N_t}{n_t}}$, where $N_t$ is the total number of simulations run from the parent node; as in the UCT algorithm). 

To this conventional tree policy, we add a habit value term, that is, the scaled likelihood of the action $a$ planned to get to a node, making its total propensity be:
\begin{equation}\label{eqn:tree_policy}
Q(a) = \dfrac{w_t}{n_t} + c * \sqrt{\dfrac{ln N_t}{n_t}} + h * p_t(a_{i}=a|a_{0}:a_{i-1})
\end{equation}
where $c$ is the exploration coefficient, here fixed to 1; $h$ is the habit coefficient; $a_i$ is the action directly leading to the node; and $a_{0}:a_{i-1}$ is the path of all preceding planned actions from the root, where $i$ indicates the current depth in the tree. Note that the win proportion and exploration terms are node-dependent, while the habit term is node-independent. Probability $p_t(a_{i}=a|a_{0}:a_{i-1})$ comes from a Bayesian non-parametric sequence model that flexibly combines the predictive power of dependencies at variable depths  \cite{teh2006hierarchical}. The model was recently used to explain motor skill learning in humans \cite{eltetHo2022tracking}). Here, we adapt it to the task of hierarchical state augmentation (HABITS) in which it provides a flexible window onto the action path leading to a state, for determining the likelihood of the next action in the search (upper right box of Figure~\ref{model_figure}). Its predictions are based on previous action sequence outputs of the planner. Formally (temporarily dropping the iteration index $t$), the model is an hierarchical Dirichlet process (HDP):

\begin{equation}\label{eqn:HDP}
a_i \sim \mathbf{G}_{\mathbf{u}_i} \sim HDP(\alpha, \mathbf{G}_{\pi(\mathbf{u}_i)})
\end{equation}

where $\mathbf{G}_{\mathbf{u}_i}$ is the vector of action probabilities given the context $\mathbf{u}_{i}$ of some previous actions before the $i$-th action; $\alpha$ is a strength parameter, controlling the resemblance to the base distribution and, therefore, the speed of sequence learning; and the base distribution $\mathbf{G}_{\pi(\mathbf{u}_i)}$, which is the probability distribution over actions given the truncated context $\pi(\mathbf{u}_{i})$ containing all but the earliest action. Applying Equation \ref{eqn:HDP} recursively performs a weighted smoothing across the action probabilities conditioned on preceding action chunks of shrinking sizes. For more details and algorithms, see \citeA{eltetHo2022tracking}. In sum, the sequence model flexibly augments the states with a \textit{weighted} window onto the previous actions, efficiently utilizing the ones that have predictive power.

Assume that the HABITS module has been trained on action sequence outputs of the planner from previous episodes (irrespectively of the quality of those plans). Then, the HABITS module interacts with the planner in two ways. It allows for biasing the one-step search towards predictable action sequence paths through the habit term in Equation \ref{eqn:tree_policy}. Moreover, by unrolling the sequence model, we also extend the node's action repertoire by action chunks (lower right box in Figure~\ref{model_figure}). Action chunks were created by sampling and attaching successor actions while the conditional entropy of the distribution over the next action was lower than the open-loop threshold parameter $\omega$. When the action chunks were considered by the tree policy, their habit value was determined by the predictability of the first action in the chunk. By choosing an action chunk, the agent jumped deeper in the tree without adding the intermediate node(s) to the tree and evaluating them, effectively stunting all the other branches that would have grown from the intervening nodes. As such, the MCTS-with-HABITS can be viewed as a hybrid between open- and closed-loop. The complexity of the chunks proposed by the HABITS scaled with the complexity of the sequences that the agent executed in past episodes, opportunistically saving as much planning cost as possible.

Our experiments tested the  unique benefits of the two separate mechanisms: the sequence bias in the one-step search, realized by the habit term in the tree policy, and the open-loop action chunks, realized by unrolling the sequence model until the open-loop entropy threshold $\omega$ is reached. Therefore, we compared three variants of our model: MCTS-with-HABITS\textsubscript{full}, utilizing both one-step biases and open-loop action chunks; MCTS-with-HABITS\textsubscript{open-loop}, utilizing no one-step bias for the actions but having the alternative of unrolling primitive actions into chunks; and MCTS-with-HABITS\textsubscript{one-step} having no action chunk alternatives but utilizing sequence predictions to bias its action choices; along with the control model, MCTS\textsubscript{vanilla}. For brevity, we will refer to MCTS-with-HABITS\textsubscript{full} as MCTS-with-HABITS. Within the scope of this paper, we did not optimize the model parameters, but selected values for demonstration (Table~\ref{model_paramsets}).

\begin{table}[!htbp]
\begin{center} 
\caption{Parameter settings for four alternative model variants. The node budget $b$, exploration coefficient $c$ and sequence learning parameter $\alpha$ were fixed across the model variants, while we varied the habit coefficient $h$ and the open-loop threshold $\omega$.} 
\label{model_paramsets} 
\vskip 0.12in
\begin{tabular}{lccccc} 
\hline
Model name                                & $b$   & $c$  & $\alpha$ &  $h$  & $\omega$   \\
\hline
MCTS-with-HABITS\textsubscript{full}      & 50    & 1    & 1        &  5    & 1.5        \\
MCTS-with-HABITS\textsubscript{open-loop} & 50    & 1    & 1        &  0    & 1.5        \\
MCTS-with-HABITS\textsubscript{one-step}  & 50    & 1    & 1        &  5    & 0          \\
MCTS\textsubscript{vanilla}               & 50    & 1    & -        &  0    & 0          \\
\hline
\end{tabular} 
\end{center} 
\end{table}

Apart from the labelled additions to the tree policy,  the MCTS algorithm was used conventionally, looping through the node selection, expansion, rollout with uniform random policy, and backpropagation of a binary value (success or failure) to the selected node and its ancestors.

\section{Results}

\subsection{Task}

\begin{figure}[!h]
\begin{center}
\includegraphics[scale=0.8]{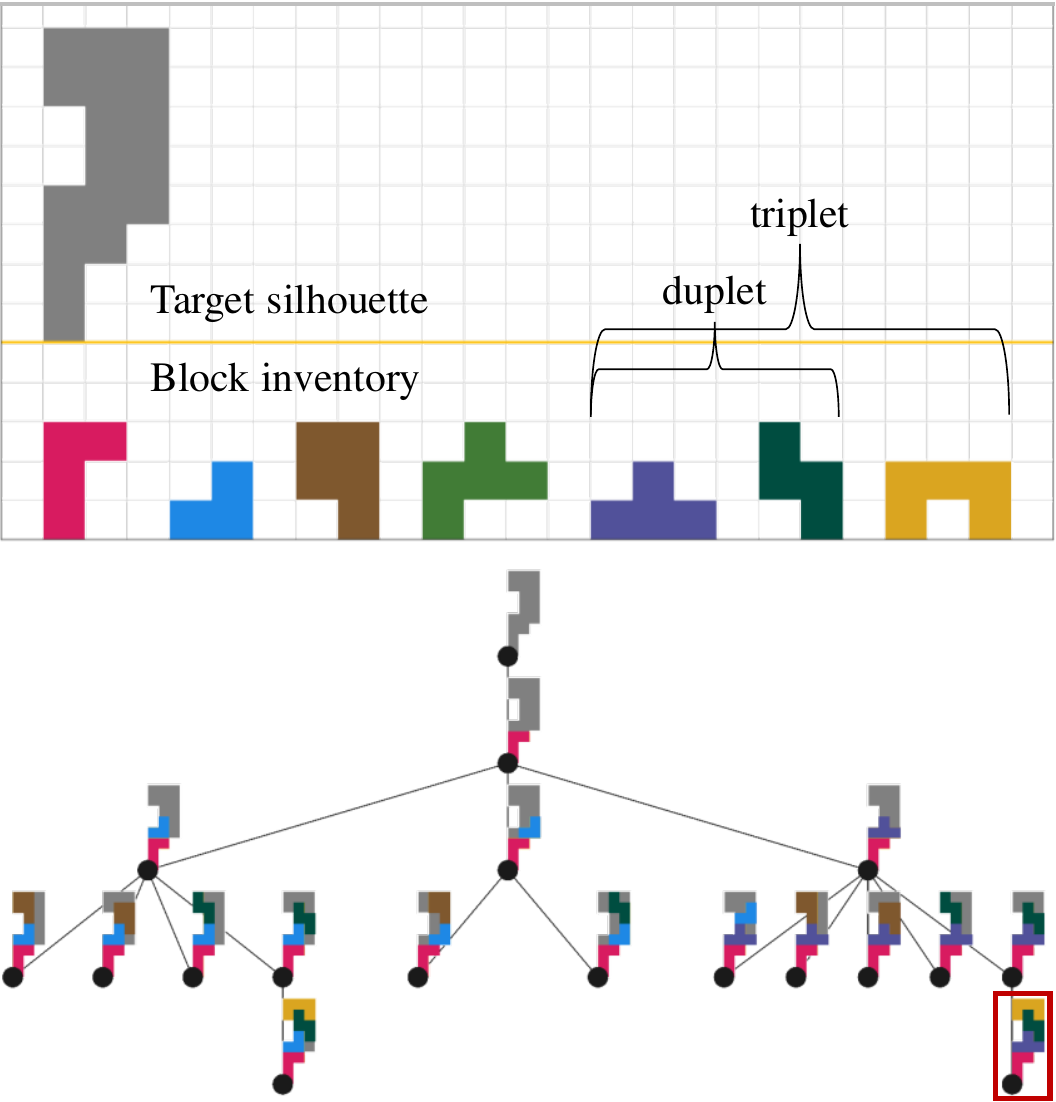}
\end{center}
\caption{(A) Schematic of an example trial of the Sticky Tangram task. The silhouette (grey area) is the target that should be built up using the building blocks (shown in different colors). The first chosen block has to be placed on the floor (yellow line); the following blocks have to be placed such that they touch at least a unit segment of a previous block's edge (hence the 'stickiness'). In the duplet condition, two blocks always occurred together, in the same relative position; in the triplet condition, three blocks formed such a chunk. Throughout the paper, these two particular chunks will be used as examples. (B) The full search tree of the example problem in (A). The leaf node in red brackets is the only solution, the other leaf nodes represent dead ends.}
\label{task_figure}
\end{figure}

We tested our model on a physical construction task inspired by the classical tangram puzzle, versions of which were recently employed by \citeA{bapst2019structured} and \citeA{mccarthy2021connecting}. Our version is called the Sticky Tangram (Figure~\ref{task_figure}A). The agent had to plan a construction that matched a target silhouette, using the seven building blocks in the inventory. The blocks could be placed into a grid space which discretized and constrained the state space. The coarse discretization is essential in order to keep the puzzle in a moderate complexity regime. The blocks had to be placed in such a sequence that they touched each other, i.e. floating blocks were not allowed.

A state included the target silhouette and the action history of the relative positions of the previously placed building blocks (Figure~\ref{task_figure}B). Adding a block to the construction constituted a primitive action; adding several blocks at once constituted an action chunk. In any state, only valid actions were considered -- that is, block placements that were inside the borders of the target silhouette and that resulted in no disjoint parts. Note that this applied to action chunks as well, meaning that actions that were unrolled into chunks that would have violated the rules of the game were discarded.

\subsection{Experiments}

\subsection{Reusing variable-size chunks}

We randomly generated silhouettes that were composed of four building blocks (Figure~\ref{task_figure}A). In the duplet condition, two building blocks formed a chunk such that they would always occur together, in the same relative position and had to be placed in a fixed order. In the triplet condition, three formed such a chunk. The unchunked blocks occurred in unpredictable relative positions. 'Chunky silhouettes' comprised the duplet and two random unchunked building blocks in the duplet condition or the triplet and one random unchunked building block in the triplet condition.

In the chunky silhouettes, the marginal probabilites of the chunked blocks were $1$, but in the unchunked blocks they were $<1$. That is because the elements of the chunks were \textit{always} present in the chunky silhouettes, per definition, but the other blocks were not. Since we wanted to study the advantage of \textit{sequence structure} in planning problems, we ensured that the overall marginal probabilities of the blocks were equal, and a chunked block was more predictable only conditioned on the other elements of the chunk. Therefore, we introduced 'random silhouettes' comprising only unchunked blocks. A 4:3 ratio of the chunky and random silhouettes in the training set ensured that the marginal probability distribution of the blocks was uniform -- that is, the agent used all the blocks equally frequently for the correct solutions.

Since the chunky silhouettes were generated using more constraints than the random ones (i.e. the chunk constraint), on average, the full search trees of chunky silhouettes are less complex. We quantified the problem complexity as the median number of nodes that the MCTS\textsubscript{vanilla} evaluated until finding the solution. Then we sampled sets of the chunky and random silhouettes such that they were matched by their complexity. We restricted the problem complexity to be $\leq50$. The example silhouette shown in Figure~\ref{task_figure}A and B has a full search tree of 18 nodes and a low complexity of 12.

\begin{figure}[!tb]
\begin{center}
\includegraphics[scale=0.87]{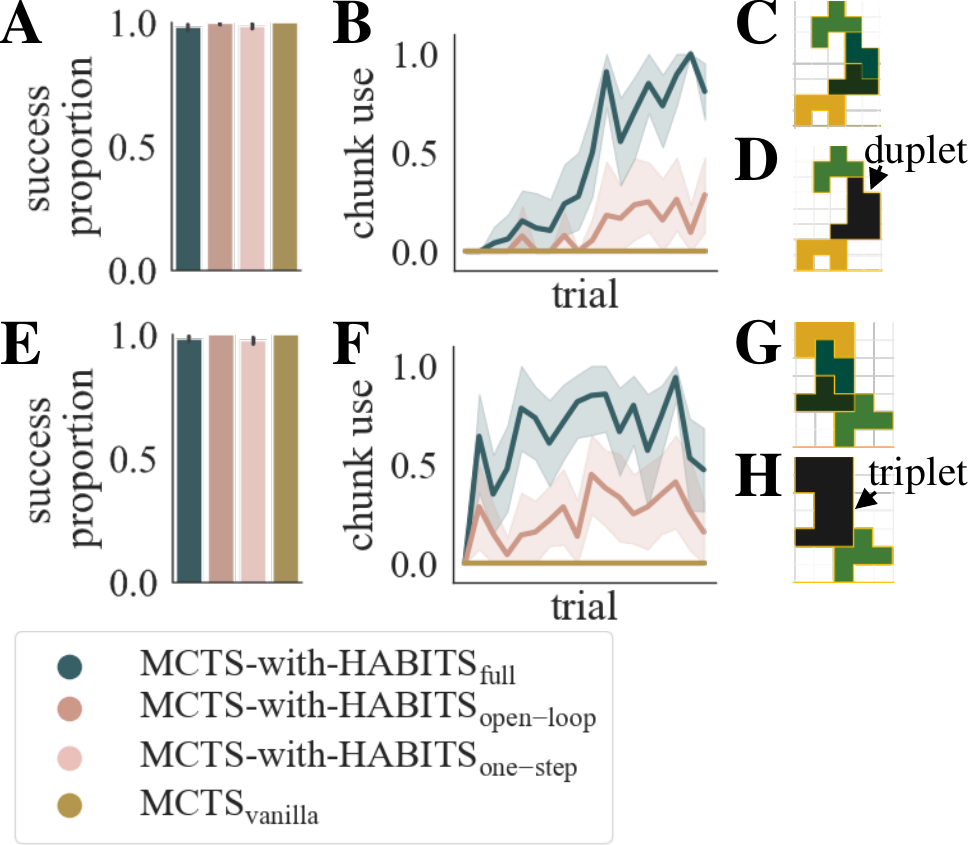}
\end{center}
\caption{Emergence of action sequence reuse in the duplet condition (A-B) and the triplet condition (C-D). Error bars and bands indicate the $95\%$ CI. (A)(E) All model variants had nearly perfect performance both in the duplet and triplet conditions. (B)(F) The two model versions that were permitted to use open-loop action chunks gradually picked up on the structure in the action sequences and became more likely to use an action chunk for the solution. In the triplet condition (F), chunk use emerged even faster than in the duplet condition (B). (C)(G) Examples of a correct solutions by the MCTS\textsubscript{vanilla}. (D)(H) Examples of solutions by the MCTS-with-HABITS model, using an action duplet (D) and an action triplet in (H). The black polygons mark chunks.} 
\label{Exp1_1}
\end{figure}

We trained four MCTS variants on 19 trials (a randomized sequence of chunky and random silhouettes). During training, all model variants used a budget of 50 nodes, enabling nearly perfect performance due to the silhouettes having complexity values $\leq50$ (Figure~\ref{Exp1_1}A and E). In the case of the three MCTS-with-HABITS model versions, the sequence module was trained on the action sequence output of the planner. We ran 32 simulations for both the duplet and triplet conditions, from different random seeds, and averaged the results.

All model variants showed nearly perfect performance given a node budget sufficient for the problems ($>96\%$ in the duplet condition and $>94\%$ in the triplet condition; Figure~\ref{Exp1_1}A and E), albeit with different patterns of chunk use. The sequence model of the MCTS-with-HABITS model versions gradually extracted the structure in the action sequences generated by the planner, that is, the action duplets and triplets in the respective conditions. The MCTS-with-HABITS and MCTS-with-HABITS\textsubscript{open-loop} models began to propose action chunks and these chunks were used more often in the correct solutions throughout the training, both in the duplet (effect of \textit{trial}: $F(1, 696)=293.70$, $p<.001$; Figure~\ref{Exp1_1}B) and in the triplet conditions (effect of \textit{trial}: $F(1, 712)=29.51$, $p<.001$; Figure~\ref{Exp1_1}F). In both conditions, the MCTS-with-HABITS was more disposed to include chunks in its plans (\textit{model} effect: $F(1, 696)=207.58$, $p<.001$ for the duplet condition; $F(1, 712)=150.39$, $p<.001$ for the triplet condition). Problems that were solved in four primitive steps by the MCTS\textsubscript{vanilla} (Figure~\ref{Exp1_1}C and G) were solved by the MCTS-with-HABITS in only three steps in the duplet condition (Figure~\ref{Exp1_1}D) and in two steps in the triplet condition (Figure~\ref{Exp1_1} H). Note that action chunks were reused in spite of the same states never reoccurring across episodes since the sequence model proposes chunks in a state-independent manner. In sum, the MCTS-with-HABITS model found and utilized the largest exploitable chunks that reoccurred in planned action sequences.

\subsection{Solving high complexity problems with less compute}

After the training trials, we gradually reduced the node budget from the initial value of 50 to 12, 8, 5, and 1. In order to compare performance across these node budgets, we froze the sequence model at the test trials. At each reduced node budget value, the agents solved four silhouettes. We ran 32 simulations for both the duplet and triplet conditions separately, from different random seeds, and averaged the results.

\begin{figure}[!t]
\begin{center}
\includegraphics[scale=0.87]{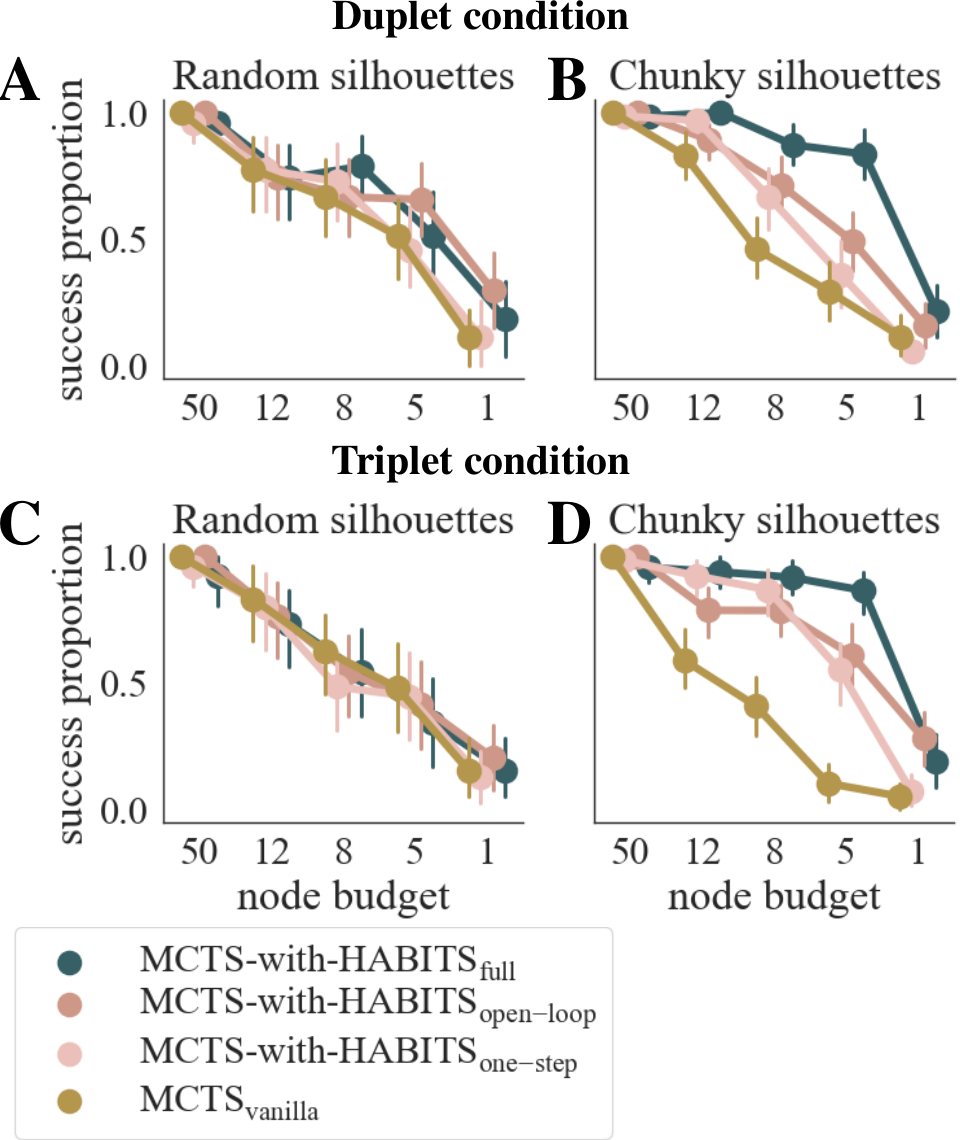}
\end{center}
\caption{Model performance in the face of node budget restriction in the test phase of Experiment 1, in the duplet condition (A-B) and in the triplet condition (C-D). Error bars indicate the $95\%$ CI. (A)(C) Performance on random silhouettes decreased for all model variants to similar degrees due to the node budget restriction. (B)(D) For chunky silhouettes, the performance of the MCTS-with-HABITS\textsubscript{open-loop} and MCTS-with-HABITS\textsubscript{one-step} was more resilient to the node budget restriction than that of the MCTS\textsubscript{vanilla}. The full MCTS-with-HABITS model showed the best performance in the face of node budget restriction.} 
\label{Exp1_2}
\end{figure}

When solving random silhouettes, all model variants' performance dropped to similar degrees as a function of node budget restriction (effect of \textit{budget}: $F(1, 1872)=124.58$, $p<.001$; effect of \textit{model}: $F(1, 1872)=0.42$, $p<.73$; Figure~\ref{Exp1_2}A and C). In the case of the chunky silhouettes, when previously learned action sequences were appropriately reusable, the resource-constrained model performance depended on whether the model was allowed to use one-step biases, action chunks, or both; this held both in the duplet (effect of \textit{model}: $F(1, 1244)=17.03$, $p<.001$; Figure~\ref{Exp1_2}B) and triplet conditions (effect of \textit{model}: $F(1, 1188)=57.16$, $p<.001$; Figure~\ref{Exp1_2}D).

\begin{figure}[!t]
\begin{center}
\includegraphics[scale=0.87]{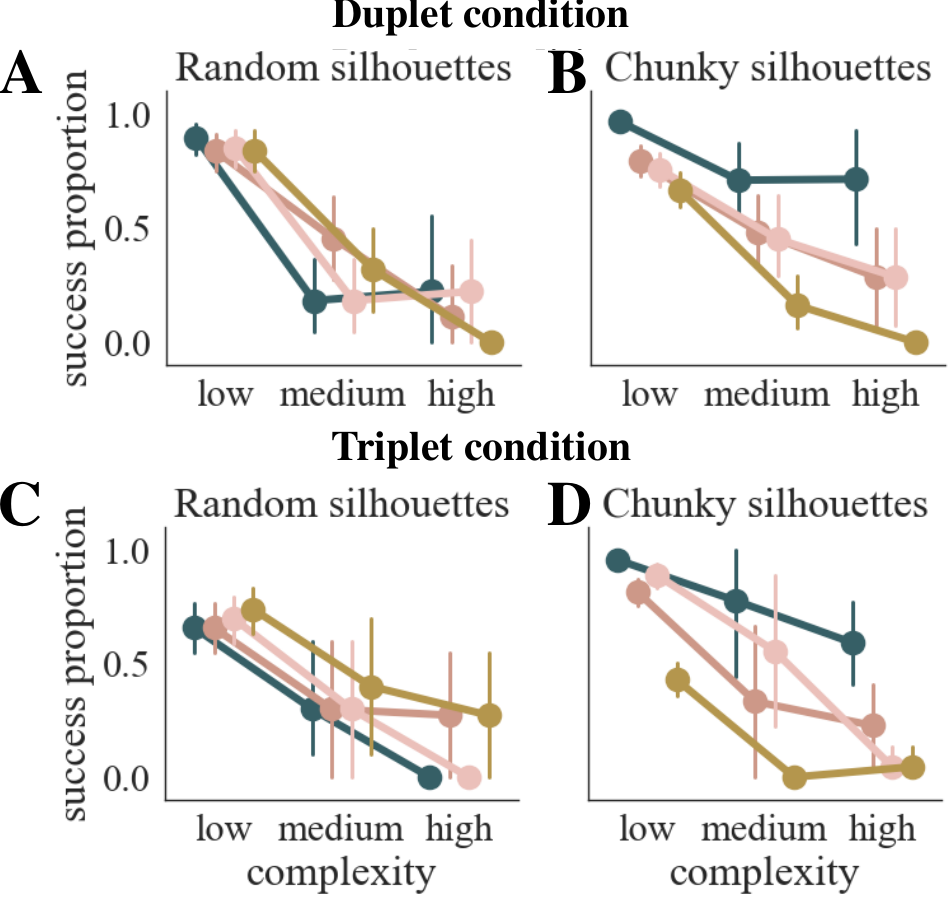}
\end{center}
\caption{Model performance as a function of problem complexity in the test phase of Experiment 1, in the duplet condition (A-B) and in the triplet condition (C-D). Error bars indicate the $95\%$ CI. Color coding is identical to that of Figure~\ref{Exp1_2}.} 
\label{Exp1_3}
\end{figure}

We then analyzed the planning performance in the resource-restricted regime of 12, 8, and 5 nodes, as a function of problem complexity. In the case of the random silhouettes, the different model variants' performance scaled with the problem complexity similarly, both in the duplet (\textit{model}*\textit{complexity} interaction: $F(6, 384)=1.72$, $p=.11$; Figure~\ref{Exp1_3}A) and triplet conditions (\textit{model}*\textit{complexity} interaction: $F(6, 364)=.52$, $p=.78$; Figure~\ref{Exp1_3}C). However, performance on the chunky silhouettes scaled differently with complexity across the model variants, in both conditions (\textit{model}*\textit{complexity} interaction in duplet condition: $F(6, 744)=1.89$, $p=.07$; Figure~\ref{Exp1_3}B; \textit{model}*\textit{complexity} interaction in triplet condition: $F(6, 764)=4.07$, $p<.001$; Figure~\ref{Exp1_3}D). This suggests that the \textit{effective} complexity of problems was reduced by relying on past action sequence statistics.

\subsection{Solving ambiguous problems}

\begin{figure}[!t]
\begin{center}
\includegraphics[scale=0.87]{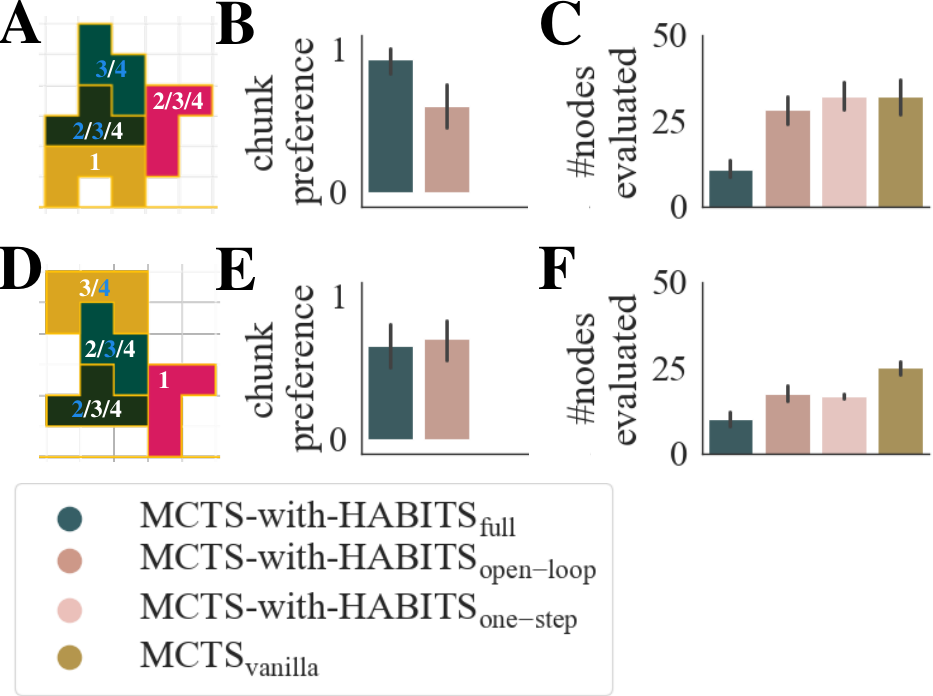}
\end{center}
\caption{Action chunk reuse in sequence-ambiguous problems in the duplet (A-C) and triplet conditions (D-F). Error bars indicate the $95\%$ CI. (A)(D) Example silhouettes whose correct solutions allowed for flexible ordering of the chunk elements. The digits indicate the possible serial positions of blocks in different solutions. The blue digits indicate the serial positions of the blocks in solutions that preserved the chunked order (same example chunks as in Figure~\ref{Exp1_1}). (B)(E) Solutions containing chunks were preferred by both model variants that were enabled to generate chunks. (C)(F) The MCTS-with-HABITS\textsubscript{full} model evaluated the least number of nodes until finding a solution.} 
\label{Exp2}
\end{figure}

We trained the model variants on chunky silhouettes that enforced an action chunk -- that is, the chunked blocks had to be placed in fixed relative positions and order for the correct solution. Here, we test the model variants on so-called 'sequence-ambiguous silhouettes' that have multiple solutions allowing for placing the chunked blocks in different orders (Figure~\ref{Exp2}A and E). Crucially, such sequence-breaking was unlikely under the sequence model, making the sequence-breaking alternative a lower-value plan.

Indeed, solutions containing chunks were preferred by both MCTS-with-HABITS and MCTS-with-HABITS\textsubscript{open-loop} models (Figure~\ref{Exp2}C and G).
We allowed for a flexible node budget and measured the number of node evaluations performed until the solution was reached (Figure~\ref{Exp2}D and H). In both the duplet and triplet conditions, the number of evaluated nodes was significantly different between the model variants, and the lowest in the case of the MCTS-with-HABITS (\textit{model} effect: $F(3, 156)=25.86$, $p<.001$ and \textit{model} effect: $F(3, 156)=42.99$, $p<.001$. To conclude, our model prefers to reuse past action sequences even when other alternatives are viable and this confers it the computational benefit of planning with smaller trees.

\section{Discussion}

Discovering the building blocks of a problem is key to its efficient solution. The behavior of animals suggest that they form action chunks that they flexibly employ for solving novel tasks. We proposed that such actions chunks can be usefully integrated into planning as well, effectively reducing the depth of problems. We introduced the MCTS-with-HABITS, a model that finds and leverages action sequence patterns at variable scales and integrates them into the planning process in order to save computational costs.

We tested the model on a physical construction task that was constrained such that certain blocks were placed in predictable positions relative to each other. 
Our method used a flexible Bayesian sequence model, which can find predictable action sequences of different sizes in past episodes, as induced by the task constraints, which it can then reuse for the planning problem at hand. We allowed the sequence model to influence the one-step search inherent to MCTS, and  also to propose multi-step node expansions to the planner. Both mechanisms gave the model an edge in performance over the baseline model when tested under resource-constraints. We showed that this was related to the model being more economical with its node budget and searching deeper on predictable paths via the multi-step expansions.

Various extensions to the method are possible. For instance, in our current algorithms, the learned action sequences only influenced MCTS' tree policy, that is, the search among the nodes whose values were evaluated. One could also use the sequence model to inform the rollout policy, potentially improving the simulation step. Equally, it would be possible to store information about the complexity of subsequent planning with nodes, and use this to adjust the use of the sequence model depending on overall demands of planning space or time. 

In the current work, the construction task that our model was tested on was more complex than typical psychology experiments but less complex than the ones many real-life applications pose. Testing the model on more complex problems and problems with partially transferable structure will be revealing about the scalability of our approach. It would be particularly interesting to look at stochastic problems, and those with intermediate rewards, which exploit more of the power of MCTS.

Our model extracts and reuses action patterns in past behavior. Another class of approaches focuses on past state visitation patterns. \citeA{mcgovern2001automatic} proposed that so-called bottleneck states can be identified at rare state transitions and that they are useful subgoals that link subproblems. Several methods based on the successor representation \cite{machado2017eigenoption, machado2021temporal} use the eigendecomposition of state succession patterns to carve the environment into subproblems. However, these approaches require the agent to have explored the entire state space extensively, a requirement that often cannot be met in machine learning and robotics. 

Applying the law of exercise to the realm of planning was inspired by the cognitive science literature. How model-free \cite{daw2005uncertainty} or value-free \cite{miller2019habits} statistics interact with model-based reasoning has been a long-standing interest in the field. Here we proposed a hybrid model that flexibly adjusts the weighting of value-based and value-free assessments by relying on predictably repeated chunks of behavior. Such sequence reuse yields a solution for optimizing the direction and depth of search \cite{sezener2019optimizing, kuperwajs2021planning}, reminiscent of humans' resource-rational strategies \cite{callaway2022rational}. Future experiments should investigate whether humans indeed use habits for planning, whether they utilize such habits of mind adaptively in order to save planning costs, and, if so, whether their planning is well described by the model presented here.

\section{Acknowledgments}


NE and PD were funded by the Max Planck Society. PD was also funded by the Alexander von Humboldt Foundation. The authors thank Tankred Saanum for helpful conversations.

\bibliographystyle{apacite}

\setlength{\bibleftmargin}{.125in}
\setlength{\bibindent}{-\bibleftmargin}

\bibliography{refs}

\end{document}